# Trust Considerations for Explainable Robots: A Human Factors Perspective

Lindsay Sanneman
Massachusetts Institute of Technology
Cambridge, Massachusetts, USA
lindsays@csail.mit.edu

Julie A. Shah
Massachusetts Institute of Technology
Cambridge, Massachusetts, USA
julie_a_shah@csail.mit.edu

## 1 INTRODUCTION

With the recent focus on explainable artificial intelligence (XAI) in the AI and robotics literature, engendering trust in increasingly complex and opaque AI and robotic systems through human-understandable explanations of system behavior has been a topic of discussion in many works [1–3, 6, 9]. As robots perform increasingly complex tasks in environments shared with human teammates, the need for explainable robot behavior will persist. Considering how explanations can impact a human teammate's trust in the robot is critical, since trust can impact reliance and use of these systems [5, 10]. Previous work in human factors has explicitly explored the importance of user trust in automated systems [8, 10]. The focus in these works is not on increasing user trust, which is often given as a motivation for XAI and explainable robots, but on appropriately calibrating trust resulting in appropriate use of these systems.

There are existing robot explanation techniques that consider trust as a success metric for the explanation system. For example, Wang et al. [12] implement a partially-observable Markov decision process (POMDP) explanation-generation technique and measure trust using survey and behavior-based data. Chakraborti et al. [4] produce robot explanations to reconcile a human's model with the robot's, and in human subject experiments, they ask participants to indicate whether they trust the robot and whether their trust increased during the study. Since human trust in robot systems can directly impact human-robot team performance, we believe that it is not only important to consider trust as a success metric for explanations but also at the explanation technique design phase.

In this paper, drawing from the human factors literature [8, 10], we suggest three important trust-related considerations for the design of explainable robot systems that can describe their behavior to human teammates: the bases of trust, trust calibration, and trust specificity. We further detail existing and potential metrics for assessing whether a person appropriately trusts a robotic system based on explanations it provides about its behavior.

## 2 BASES OF TRUST: PURPOSE, PROCESS, AND PERFORMANCE

Lee and See [8] introduce purpose, process, and performance as three bases for user trust in an automated system. They assert that providing users with information about these elements can help to ensure appropriately calibrated trust. We believe that it is important for robots to communicate information about all three elements through explanations about their behavior.

Lee and See [8] define "purpose" as the degree to which the system is being used within the realm of the designer's intent. A robot providing a purpose-related explanation might tell a human the tasks it can and cannot do according to its design. For example, a manufacturing robot could say to a human teammate, "I can lift parts weighing up to 10kg and move them, but I cannot sense if you are in my workspace." "Process" refers to the appropriateness of a system's algorithm for the situation in which it is working and the extent to which it can contribute to the team's goals. A robot providing a process-related explanation might give information about how it performs its tasks. For example, the manufacturing robot could tell its human teammate, "I decide which objects to move and where to move them based on pre-programmed schedules and cannot move un-programmed objects." Finally, "performance" is related to an automated system's demonstrated operations, including characteristics such as reliability, predictability, and ability. A robot providing performance-related information in an explanation could explain its specifications, limitations, or confidence levels. For example, the manufacturing robot could say to its human teammate, "I correctly place items I move 90% of the time."

There are two important aspects of performance-related communication provided by robots: communication about the robot's task performance and communication about its explanation performance. For example, a robot might perform a camera-based object detection and classification task very well and accurately report its classification accuracy. However, it might not perform as well at explaining the factors that impacted its classification and edge cases in which it might make mistakes. Information about both how a robot performs at decision-making processes and at explanation production can help users understand the trustworthiness of both the robot (with regard to its given tasks) and its explanations.

## 3 TRUST CALIBRATION

Lee and See [8] define "trust in automation" as the attitude that an agent will help achieve an individual's goals in a situation characterized by uncertainty and vulnerability. They emphasize that it is only appropriate to increase trust when the system is trustworthy. In other words, someone should only maintain the attitude that an agent will help them achieve their goals to the extent that the agent is actually able to do so. More important than increasing trust in robots, as is stated as a goal in [2], is appropriately calibrating trust to ensure appropriate use [8]. Overtrust in a system can lead to overuse, while undertrust can lead to disuse [10]. Thus, when developing explainable robots and considering how explanations influence trust, the goal should not be to increase trust but to align user trust with the system's true capabilities or purposes. Appropriate trust calibration, enabled through robot explanations, can be achieved through the robot sharing information about its capabilities, limitations, and confidence, as discussed in section 2.

## 4 TRUST SPECIFICITY: LOCAL VERSUS GLOBAL TRUST CALIBRATION

Lee and See [8] discuss the concept of specificity of trust in automation. They define "functional specificity" as the differentiation of trust between functions, subfunctions, and modes of automation. They argue that with a high degree of functional specificity, a person's trust can reflect the capabilities of specific subfunctions and modes of an automated system. Explainable robot systems must be able to provide information to support functionally specific trust in robot systems, which we refer to as "local trust" in the system.

Local trust calibration is important for robot systems, because a single robot may not be uniformly trustable across all contexts. Consider a semi-autonomous robotic vehicle which possesses extensive training data collected from highway driving but relatively little training data from urban environments. This vehicle might perform adequately without human assistance on highways but worse on city streets. In such a scenario, an explainable robot system must be able to support user understanding of its differing performance across the various contexts. Developing explainable robot systems that are able to support local trust calibration can also be beneficial for development of user-specific explanations, tailored to the needs of different individuals who work with the robot.

While local trust calibration can help different users appropriately trust and engage with a robot in different contexts, overall understanding of a robot's performance and abilities is important for allowing users to determine how they might trust the robot in new scenarios. In our semi-autonomous vehicle example, while local trust can be calibrated for different driving contexts, it is likely not possible for a user to encounter every contextual scenario while learning about how the vehicle operates. Therefore, some global information about the system's overall performance (such as number of miles between vehicle failures) could help users to maintain a prior on how trustworthy the system will be in new scenarios. In sum, to support global trust calibration, an explainable robot system should provide information about the robot's global performance towards its overall goal in order to improve users' understanding of the robot's overall trustworthiness.

## 5 METRICS FOR EXPLAINABLE ROBOTS

Some existing literature assesses user trust in AI or robotic systems primarily through survey questions. For example, Chakraborti et al. [4] ask users to rank whether they trusted the robot in their setup to work on its own and whether their trust in the robot increased over the course of the study (on a Likert scale). Similarly, Wang et al. [11] ask users to rate their trust in a system on a Likert scale. Wang et al. [12] also ask human users to rank their trust in a robot system under different conditions, and they additionally use compliance, defined as the number of participant decisions that match the robot's, as a behavior-based metric for measuring trust in the system.

In the human factors literature, Dzindolet et al. [5] explore the role of trust in automation reliance and conclude that trust impacts reliance upon automated systems regardless of a system's actual ability, suggesting that reliance could be an appropriate proxy metric for assessing how explanations impact user trust. In the XAI literature, Hoffman et al. [7] emphasize the importance of trust calibration and suggest that a trust scale should ask two primary questions: whether a user trusts the system's outputs and whether they would follow its advice. They further propose a trust scale that assesses the value of XAI systems based on the user's trust in the AI system before and after it provides explanations. We recommend using a trust scale such as this in conjunction with a behavior-based metric (such as reliance, compliance, etc.) in order to determine whether humans appropriately trust and use robot systems in response to provided explanations. Ideally, a user's trust and reliance should correspond to a robot's capabilities.

## 6 CONCLUSION

Appropriately calibrated trust in robotic systems is critical to human-robot team performance and can be enabled through robot explanations of their behavior. Human trust in robots can be supported through robot explanations about their purpose, process, and performance. Trust should also be appropriately calibrated at both a local level (functional specificity) and a global level. Finally, in measuring the success of robot explanation techniques, researchers should measure the impact of explanations on human trust using a combination of a trust scale and a behavioral metric, such as reliance or compliance.